\documentclass{OAGM}
\OAGMarXiv{1304.1876}

\usepackage{amssymb}
\usepackage{amsmath}
\usepackage{abbrevs}
\usepackage{graphicx}
\usepackage{color}
\usepackage{multirow}
\usepackage{balance}
\usepackage{fmtcount}
\usepackage{url}
\newname\eg{e.g.}
\newname\Eg{E.g.}
\newname\etc{etc.}
\newname\ie{i.e.}
\newname\etal{\em et al.}
\newname\cf{cf.}
\newname\Fig{Figure} % Fig.
\newname\Eq{Eq.}
\newname\Sec{Section} % Sec.
\newname\Tbl{Table} % Tab.

\def\reffig#1{\Fig~\ref{#1}}
\def\reftbl#1{\Tbl~\ref{#1}}

\def\mysection#1{\vskip 0.1\baselineskip\noindent{\bf #1.}}
\newcommand{\qm}[1]{``#1''}

\def\sizefigOne{0.95} % 0.95
\def\sizefigOneSmall{0.95} % 0.75
\def\sizefigTwo{0.465} % 0.48

\def\vspacefigPre{0mm} % before a figure
\def\vspacefig{0mm} % between figure and caption

\def\vspacefigPost{0mm}
\def\vspacetablePre{0mm}

\def\vspacetab{0mm}

\definecolor{markcolor}{rgb}{1,0,0}

\graphicspath{{figures/}}

\title{Counting people from above: Airborne video based crowd analysis}
\author{Roland Perko, Thomas Schnabel, Gerald Fritz, Alexander Almer, Lucas Paletta\\
JOANNEUM RESEARCH Forschungsgesellschaft mbH\\
DIGITAL - Institute for Information and Communication Technologies\\
Remote Sensing and Geoinformation, Steyrergasse 17, 8010 Graz, Austria}

\begin{document}

%\mainmatter  % start of an individual contribution


%\titlerunning{Airborne based Crowd Monitoring for Security Applications}


%
%\authorrunning{Airborne based Crowd Monitoring for Security Applications}

%\institute{JOANNEUM RESEARCH Forschungsgesellschaft mbH\\
%DIGITAL - Institute for Information and Communication Technologies\\
%Remote Sensing and Geoinformation, Steyrergasse 17, 8010 Graz, Austria\\
%\mailsa\\%\hspace{1cm}
%\url{http://www.joanneum.at}}
%\toctitle{Lecture Notes in Computer Science}
%\tocauthor{Authors' Instructions}
\maketitle
%
% ---------------------------------------------------------------------------------------------------------------------
\begin{abstract}%\vspace{-2mm}
Crowd monitoring and analysis in mass events are highly important technologies to support the security of attending persons. Proposed methods based on terrestrial or airborne image/video data often fail in achieving sufficiently accurate results to guarantee a robust service. We present a novel framework for estimating human count, density and motion from video data based on custom tailored object detection techniques, a regression based density estimate and a total variation based optical flow extraction. From the gathered features we present a detailed accuracy analysis versus ground truth measurements. In addition, all information is projected into world coordinates to enable a direct integration with existing geo-information systems. The resulting human counts demonstrate a mean error of 4\% to 9\% and thus represent a most efficient measure that can be robustly applied in security critical services. 
\end{abstract}

% ---------------------------------------------------------------------------------------------------------------------
\section{Introduction}

The recognition of critical situations in crowded scenes is very important to prevent escalations and human casualties. On large scale events, like music festivals or sport events, important parameters for estimating the riskiness of a situation are, as follows, the number of persons, the density of individuals per square meter, the general motion direction of groups of people and motion patterns (like dangerous forward and backwards motions in front of a stage or an entrance). These parameters can be used to estimate the \textit{human pressure} which indicates potential locations of violent crowd dynamics \cite{helbing2007dynamics}. Despite the huge number of security forces and crowd control efforts, hundreds of lives are lost in crowd disasters each year (like at Roskilde Festival in 2000, or in Mina/Makkah during the Hajj in 2006, or in Duisburg at Love Parade in 2010). In the future, the presented framework will provide sufficiently robust cues to prevent such disastrous incidences. 

In this paper we introduce a setup based on HD video data which can either be captured from a tower-mounted camera or from an airborne vehicle (airplane, helicopter, UAV). The resulting video, capturing parts of the crowded scene, is analyzed with computer vision techniques which extract the target parameters (count, density, motion). To be able to pipe such information in a crowd simulation framework the \textit{per-pixel} information has to be geo-referenced into a world-coordinate system. This enables to measure in physical units, \eg number of persons per square meter and motion in meters per second.
%, instead of number of persons per pixel.
A crucial parameter to detect critical situations in humans crowds is the \textit{human pressure} $P$, defined by $P(\boldsymbol{x},t)=\rho(\boldsymbol{x},t)\textrm{Var}(V(\boldsymbol{x},t))$ where $\boldsymbol{x}$ is the spatial location, $t$ the time, $\rho$ the estimated density and $V$ the motion \cite{helbing2007dynamics}, which can be estimated employing the proposed framework.
Such information can then be used to alert security staff who then triggers appropriate actions, like opening or closing a gate or restricting the access of following people.
%
%
%In future the presented workflow could be integrated in a surveillance system which will help to prevent disastrous incidences, like at Roskilde Festival in 2000, or in Mina/Makkah during the Hajj in 2006, or at Love Parade in 2010.
%
%Since it is very important to preserve the privacy of each individual, a workflow is proposed that uses the original video data only to extract image features which do not hold any detailed information on  individuals. In a real time scenario the original video frames can be directly discarded and never stored on any device (or significantly downscaled). Thus, the privacy of the attending people would not be intruded.
%
%Crowd monitoring is an important task, especially for surveillance and for security planning. In case of big events, like music festivals or sport events, and in general when many persons are
%in a crowed scenario, it becomes crucial to know how many persons are at which location and moving in which direction. Such information can be used to trigger security events, like opening or %closing of a gate, restricting the access of following people, \etc.
%Avoid dangerous situations (like Roskilde Festival 2000, Love Parade 2010)
%
%TBD\\
%Important parameters: Number of persons, density, direction of movement.
%
\mysection{Our contribution} The main difference in our approach to the related work is to apply higher order features for density estimation and provide an accurate performance analysis in a geo-referenced framework, such as, using an object detector tailored for person detection, learning the density estimate from image features w.r.t. a given ground truth (can be seen as an automatic feature selection) and rectifying all information from 2D image geometry to 3D world coordinates. In addition, the proposed framework is general and could be combined with any existing visual features, with any object category and with any object detection method. For example, it could be applied - appropriate features presumed - to count trees or cars in airborne videos.
%
%The main difference in our approach to the related work is to apply higher order features for density estimation and provide an accurate performance analysis in a geo-referenced framework, such as (1) using an object detector tailored for person detection, (2) learning the density estimate from image features w.r.t. a given ground truth (can be seen as an automatic feature selection), (3) rectifying all information from image geometry to world coordinates and (4) providing a detailed accuracy analysis of the approach. In addition, the proposed framework is general and could be combined with any existing visual features, with any object category and with any object detection method. For example, it could be applied - appropriate features presumed - to count trees or cars in airborne videos.
%
%
% ---------------------------------------------------------------------------------------------------------------------
\section{State of the art}
Some principles for crowd monitoring and person counting have been published. For example, \cite{chan2008privacy} count people in an outdoor scenario based on a fixed mounted static video camera using a motion segmentation followed by a feature extraction that serves as input for a Gaussian regression model. The main drawback w.r.t. our application is the prior motion segmentation. Such a system can only identify moving people, therefore all standing people are not counted. In addition, other moving objects like cars or pets will also appear in the motion segmentation.
Authors of \cite{butenuth2011integrating} detect individual people and crowd outlines from airborne nadir looking images. While isolated persons are detected using a custom tailored object detector, regions containing crowds are recognized when many local features (\textit{features from accelerated segment test (FAST)}) jointly occur. The work does not contain an accuracy analysis and lacks a concept of how to map potential crowd regions to estimated person counts. It also seem problematic to define regions of crowds by low-level features, as in an arbitrary scenario also other objects than people will give a high FAST response (like \eg textured vegetated areas).
The work of \cite{sirmacek2011automatic} also deals with airborne nadir looking images. This very interesting approach is similar to our methodology in terms that it extracts local features (in this case again FAST) and uses them to estimate the crowd density. The authors also include a feature selection step to reject local features which potentially are not corresponding to persons. The density itself is extracted using a kernel density estimate based on the feature occurrence. The number of individuals is spatially aggregated also using the FAST responses.

In the following we discuss related work in particular for object counting, density estimation, motion estimation and geo-referencing.
%
% -----
%
\mysection{Object Counting and Density Estimation} There are three main methodologies:
\textit{(1) Counting by detection:} The idea is to detect each individual object instance in the image and count their number (actually this is how human count). However, in computer
vision object detection is far from being solved \cite{everingham2009the} and the detection is a harder problem than counting alone. Huge problems arise when objects are overlapping and occlude each other.
\textit{(2) Counting by regression:} Those methods try to find a mapping from various image features to the number of objects using supervised machine learning methods. However, those methods
do not use the location of the objects in the image instead they just find the regression to a single number, \ie the number of objects. Therefore, huge training datasets are necessary
to achieve useful results \cite{kong2006a}.
\textit{(3) Counting by density estimation:} The main concept is to estimate an object density function whose integral over any image region gives the count of objects within this region \cite{lempitsky2010learning}. For learning the proposed methods employ the ground truth location of objects and the learning can be posed as a convex linear or quadratic program. An additional benefit of the method is that after learning the density function can be estimated by simple multiplication of the individual features with learned weights and is therefore very efficient.
%
% -----
%
\mysection{Motion Estimation} Estimating small motions from adjacent video frames is considered to be solved, or to state it differently, the accuracy of state-of-the-art algorithms are sufficient for our needs. The so-called optical flow can be extracted by total variation methods in image geometry, \eg \cite{zach2007a}.
%The real flow can be extracted by geo-referencing the flow vectors using both image models which yield flow in metric units, in particular in $[m/s]$.
%
% -----
%
\mysection{Geo-Referencing} Geo-referencing, also called ortho-rectification, is a standard method in photogrammetry and in remote sensing (\cf \eg \cite{kraus2007photogrammetry}) which projects the image onto the earth's surface in a given map projection. To be able to handle the distortions due to the topography a digital surface model (DSM) is used (global digital surface models like SRTM\footnote{\url{http://srtm.csi.cgiar.org}} or ASTER GDEM\footnote{\url{http://gdem.ersdac.jspacesystems.or.jp}} are freely available). If the terrain is rather flat the DSM can be replaced by the knowledge of the mean terrain height. For areas containing many obstacles like stages, bridges, etc. a laser scanner model will deliver most accurate results.

%
%
% ---------------------------------------------------------------------------------------------------------------------
\section{Methods}
\subsection{Workflow}
\begin{figure}[!h]
\vspace{\vspacefigPre}
\centering
\resizebox*{\sizefigOneSmall\columnwidth}{!}{\includegraphics{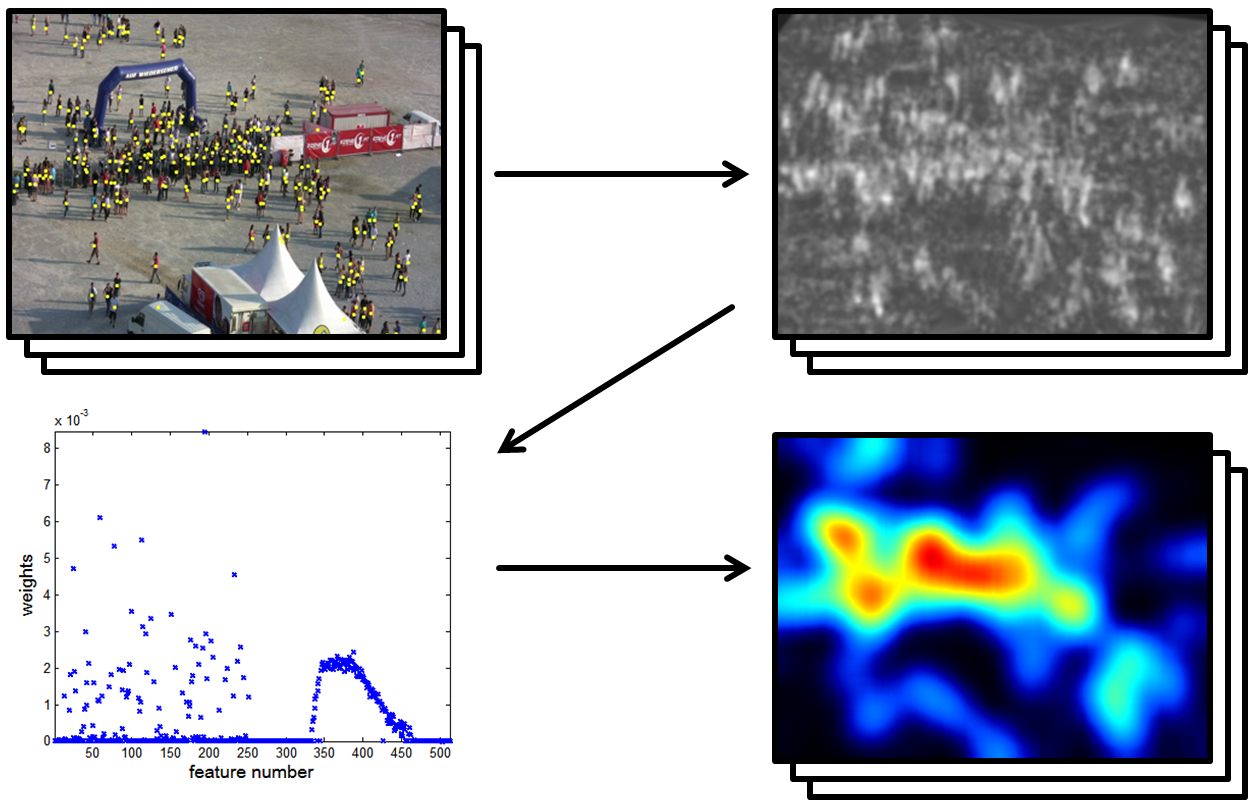}}
\vspace{\vspacefig}
\caption[]{Our proposed workflow for human density estimation: An image with annotated humans (yellow dots), discretized features (in this specific case the results of an object detector), the learned weights for each feature and the estimated human density function (estimated count equals 250) are shown.} \label{figure:WF_density}
\vspace{\vspacefigPost}
\end{figure}

\begin{figure}[!h]
\vspace{\vspacefigPre}
\centering
\begin{tabular}{cc}
\resizebox*{\sizefigTwo\columnwidth}{!}{\includegraphics{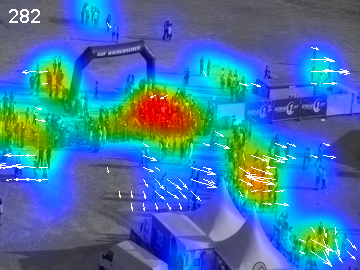}} & \resizebox*{\sizefigTwo\columnwidth}{!}{\includegraphics{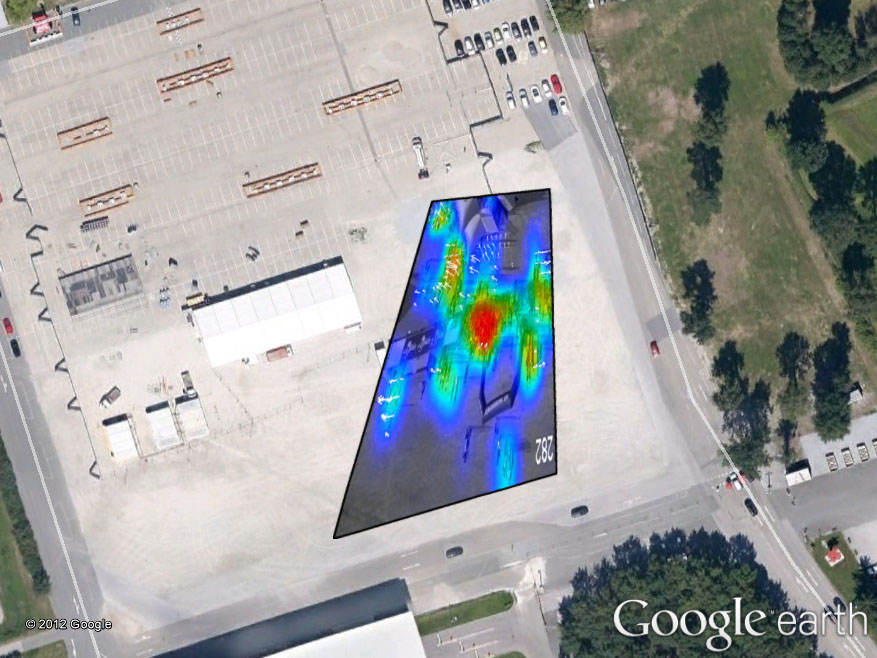}}
%(a) & (b)\\
\end{tabular}
\vspace{\vspacefig}
\caption[Geo]{Geo-referencing of a given image, the human density and motion estimate for test site \textit{Lakeside}: (left) input image with superimposed color coded human density function, motion, and estimated number of individuals and (right) the geo-referenced version of (left) shown as Google Earth$^{\decimal{footnote}}$ overlay.} \label{figure:density_image_geo}
\vspace{\vspacefigPost}
\end{figure}

The proposed approach is sketched in \reffig{figure:WF_density} and in \reffig{figure:density_image_geo}. The main idea is to extract image features which are related to the human density by machine learning techniques. We employ discretized features where the learning provides a weight for each feature number. Thus, after learning the density function can be calculated by simple multiplications. In addition, the density estimate is a real density function, meaning that the integral over the density yields the object count (therefore, the integral over a subregion holds the number of objects in this particular region). The motion between video frames is extracted using a variational method. All gathered information is then geo-referenced and can therefore be visualized and processed in any geographic information system. \reffig{figure:density_image_geo} shows a video frame superimposed with the estimated density and motion and the same information geo-referenced and overlayed in Google Earth.
\addtocounter{footnote}{1}

\subsection{Object Counting and Density Estimation}
For object counting and density estimation we employ the method by \cite{lempitsky2010learning}. This method takes dense discretized feature maps extracted from the input images and learns the density estimate via a regression to a ground truth density. Thus, each pixel has to be described by a feature vector of the following form $f=(0,0,\dots,0,1,0,\dots,0)$ which is 1 at the dimension of the corresponding discretized feature and otherwise 0. Since we want to detect persons we apply the object detector of \cite{felzenszwalb2010object} with the learned model for persons of the VOC 2009 challenge \cite{everingham2009the}. This detector yields confidence values which have to be discretized. As we know from experience and previous tests that very small and very high confidences are useless for object counting, we set the minimal value to $-4.0$ and the maximal to $-0.6$ for all tests. High confidences usually only occur on isolated non-occluded persons, \ie not in crowds. If we would not saturate the confidences, the density estimation would put too much emphasize on such objects. These bounds are used to scale the confidences to $[0,255] \in \mathbb N$. Now, each of the possible 256 values define a feature vector, as discusses above, which is 1 at the position of the confidence value. Therefore, it yields 256 individual features (\cf \reffig{figure:WF_density}).
\footnotetext[\value{footnote}]{\url{http://earth.google.com}}
In addition, we extract dense scale-invariant feature transform (SIFT) descriptors \cite{lowe2004distinctive} using the implementation in \cite{vedaldi08vlfeat} for each pixel. To be able to discretize this information we take 256 SIFT prototypes \cite{lempitsky2010learning} and the closest prototype for each descriptor defines the quantized SIFT number. Therefore, for each pixel we get a discretized SIFT value in $[0,255] \in \mathbb N$. These additional 256 features are employed to test if simpler cues than object detector confidences could yield useful results. For evaluation we train the density estimation framework for each feature class individually and for both, which is done by stacking the features.

The training itself minimizes the regularized \textit{Maximum Excess over SubArrays (MESA)} distance where we use the $L_1$ and the Tikhonov regularization \cite{tikhonov1977solutions} to solve the linear or quadratic equation system (\ie $\min_{x}{||Ax-b||}$ or $\min_{x}{||Ax-b||+||(x' \Gamma x)/2||}$ with $||x \ge 0||$ and Tikhonov matrix $\Gamma$ being the identity matrix in our case). All details of this methodology are given in \cite{lempitsky2010learning}. The result is a weight for each of the discretized features and the resulting density is calculated by multiplying the according weight with the extracted feature value. Thus, for each pixel the density function is given and the sum over all pixels represents the number of objects in the image, \ie our person count.

Therefore, in the testing phase the discretized features are extracted for each image and multiplied by the learned weight vector directly resulting in the density estimation per pixel and corresponding person count. It should be noted that this approach introduces virtually no overhead over feature extraction \cite{lempitsky2010learning}. In case of very efficient feature extraction methods, like decision tree and forests \cite{sharp2008implementing} or cascades of boosted weak classifiers \cite{viola01rapid}, the whole density estimation would also run in real time.
\subsection{Motion Estimation}
The motion is estimated based on the optical flow in image geometry \cite{zach2007a} where we used the implementation at\footnote{\url{http://www.gpu4vision.org}}. To get a more robust estimate the flow is not gathered from two adjacent video frames but from frames with a temporal distance of 10 frames. In addition a given number of those flows are temporally averaged to ensure smooth motion vectors.
\subsection{Geo-Referencing}
To keep it simple we define a common map frame for each of our test sites in WGS84 UTM 33 North projection (EPSG 32633) since our sites are located in western Austria, Europe. Then for each image and for each column/line coordinate the according 3D world coordinate is calculated which are used to rectify the density and motion information.
\mysection{Density} For geo-referencing the density we project each density pixel into the common frame. If a pixel gets hit more than once the values are summed up. This ensures that the sum of the density, \ie the human count, stays the same in image and world coordinates. Since it happens that some pixels are hit more often than their neighbors due to aliasing, the whole geo-referenced density is smoothed using a Gaussian kernel.
\mysection{Motion} In image geometry we cannot differentiate between object motion and camera motion.
% However, when transforming the reference image coordinate into the common frame using the reference transformation and the corresponding matched image coordinate with the search transformation, absolute world coordinates can be extracted. These two world coordinates define the real object motion independent of the camera movement.
Therefore, we transform the reference 2D image coordinate and the according search 2D images coordinate (\ie gathered via optical flow) into 3D world coordinates. These two world coordinates define the real object motion vector independent of the camera movement. Since the temporal difference of the two input video frames is know, the speed of the motion can be calculate in meters per second.
%
%
%
% ---------------------------------------------------------------------------------------------------------------------
\section{Results}

\subsection{Test Data}
\begin{figure}[!h]
%\vspace{\vspacefigPre}
\centering
\begin{tabular}{cc}
\resizebox*{\sizefigTwo\columnwidth}{!}{\includegraphics{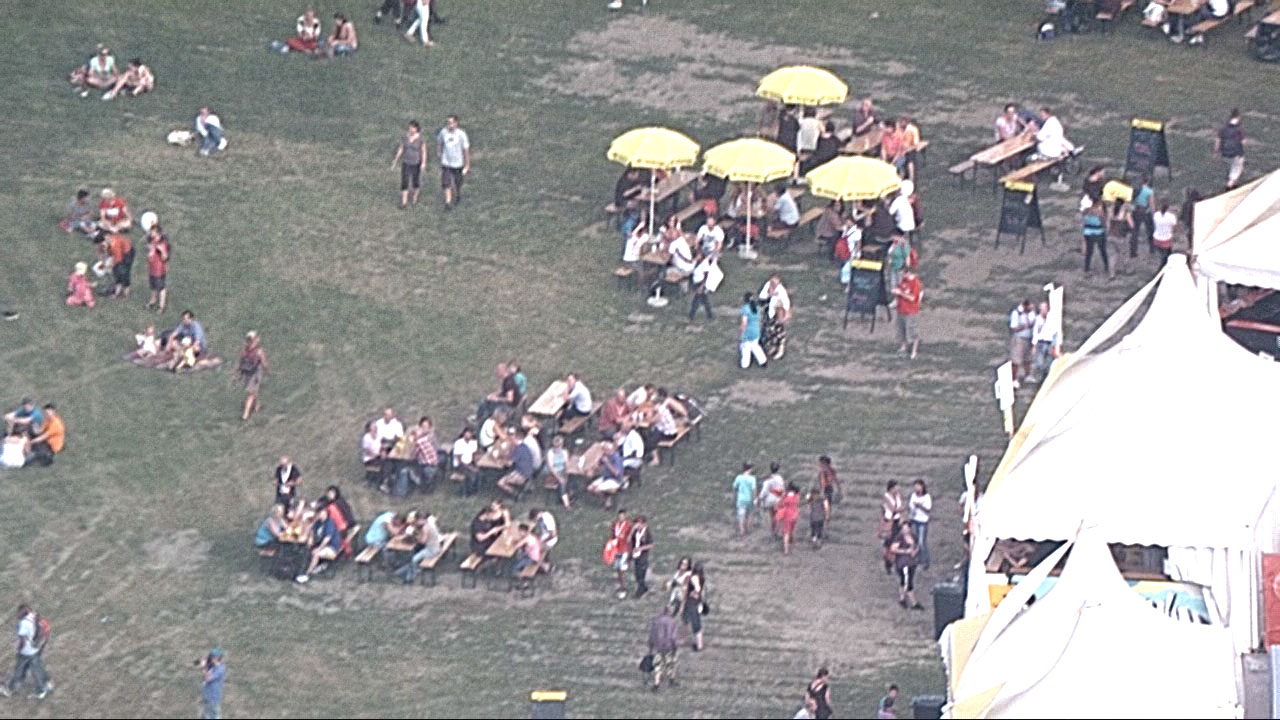}} & \resizebox*{\sizefigTwo\columnwidth}{!}{\includegraphics{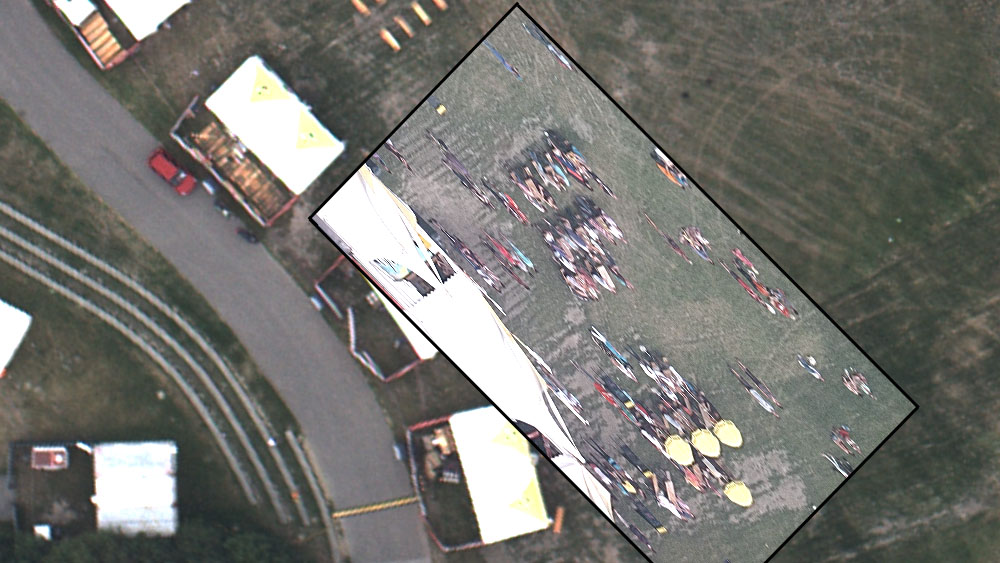}}
\end{tabular}
\vspace{\vspacefig}
\caption[Geo]{Geo-referencing of a given image for the test site \textit{Donauinsel}: (left) Airborne video frame and (right) the geo-referenced version of (left) overlayed on a true ortho image with 4cm GSD.} \label{figure:density_image_geo_DI}
%\vspace{\vspacefigPost}
\end{figure}

For evaluation of the presented concept videos from two different scenarios were acquired in HD quality. The first one, referred as \textit{Lakeside}, originates from a music festival in Styria, Austria (\cf \reffig{figure:density_image_geo}). The video camera was mounted on a tower (approximately 30 meters above ground). The camera was therefore more or less static with small jiggling due to wind. To geo-reference the scene only one image was manually rectified and defines the geometry for all other images. The second scenario, called \textit{Donauinsel}, originates from a huge open air festival in Vienna, Austria (\cf \reffig{figure:density_image_geo_DI}). Here the video camera was mounted on an airplane. For geo-referencing, the meta-data (GPS/IMU) supplied by the camera system was taken for each frame. Since every frame has a different exterior parameters, it was necessary to geo-reference every frame independently. \reftbl{table:test_data} lists the details of the video setups and parameters.
We also manually labeled many frames to get the ground truth person counts in training and later in the testing phase (overall over 23500 persons were annotated with a mean height of 90 pixels, \cf \reftbl{table:test_labels}). It is important to note that the scenes for learning are similar however different than the testing scenes. Since the \textit{Lakeside} scenario contains a much larger data set, most of the experimental results are focused on this set. The \textit{Donauinsel} scenario contains insufficient images for sustainable training and testing. In addition, the density estimate is evaluated in detail since the motion estimation can be solved by state-of-the-art algorithms.
\addtocounter{footnote}{1}
\footnotetext[\value{footnote}]{\url{http://www.diamond-air.at}}
%
%\addtocounter{footnote}{1}
%\footnotetext[\value{footnote}]{\url{http://www.diamond-air.at}}

\begin{table}[!h]
\vspace{\vspacetablePre}
\centering
\scriptsize{
\begin{tabular}{|c|c|c|c|c|c|}
\hline            & Image size & Frame & $\!$Number of$\!$ & $\!$Length$\!$     & Camera parameters \\
                  & in pixels  & rate  & $\!$frames$\!$    & $\!$in m:ss$\!$    & \\
\hline
\hline $\!$\textit{Lakeside}$\!$   & $\!$$1440 \times 1080$$\!$ & 25 & 6801 & 4:32 & $\!$Canon HV30 camera$\!$ \\
                  &                    &    &      &      & $\!$fixed mounted on a tower$\!$ \\
\hline $\!$\textit{Donau-}$\!$ & $\!$$1280 \times 720$$\!$  & 50 & 721  & 0:14 & $\!$FLIR Star Safire HD camera$\!$ \\
           \textit{insel}       &                    &    &      &      & $\!$mounted on DA42 MPP  airplane$^{\decimal{footnote}}$ \\%\footnotemark$\!$ \\ %$^\decimal{footnote}}$ \\
\hline
\end{tabular}}
\vspace{\vspacetab}
\caption[Test data sets.]{Test video data sets for the two scenarios.} \label{table:test_data}
%\vspace{\vspacetabPost}
\end{table}

\begin{table}[!h]
\vspace{\vspacetablePre}
\centering
\scriptsize{
\begin{tabular}{ccc}

\begin{tabular}{|c|c|c|c|c|}
\hline $\!$\textit{Lakeside}$\!$   & $\!$nr. of$\!$ & \multicolumn{3}{|c|}{persons}\\
                  & $\!$images$\!$    & total & mean &  std \\
\hline $\!$Training$\!$   & 12        &  3154 & 263  &  7.3 \\
\hline $\!$Testing$\!$    & 68        & 18884 & 278  & 13.2 \\
\hline
\end{tabular}
&
$\!$$\!$$\!$$\!$$\!$$\!$
&
\begin{tabular}{|c|c|c|c|c|}
\hline \textit{Donau-} & $\!$nr. of$\!$ & \multicolumn{3}{|c|}{persons}\\
       \textit{insel}  & $\!$images$\!$    & total & mean & std \\
\hline $\!$Training$\!$   & 5         & 672   & 134  & 41.7 \\
\hline $\!$Testing$\!$    & 6         & 848   & 141  & 35.8 \\
\hline
\end{tabular}
\end{tabular}}
\vspace{\vspacetab}
\caption[Test data.]{Manually labeled persons for the two scenarios.} \label{table:test_labels}
\end{table}
\subsection{Density Estimation}
\begin{figure}[h]
%\vspace{\vspacefigPre}
\centering
\begin{tabular}{c}
\resizebox*{\sizefigOne\columnwidth}{!}{\includegraphics{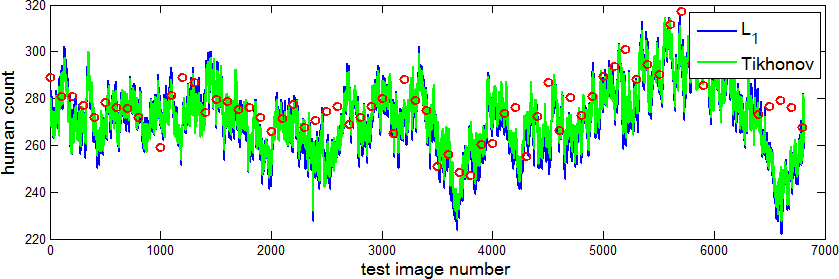}}\\
\end{tabular}
\vspace{\vspacefig}
\caption[]{Person counting: Estimated person count using $L_1$ regularization (blue) and Tikhonov regularization (green) for the \textit{Lakeside} scenario. The red dots indicate the manually measured ground truth for the test images.} \label{figure:counts}
\vspace{\vspacefigPost}
\end{figure}

\mysection{Learning} The accuracy of the learning process is listed in \reftbl{table:accuracy}. It can be seen that the used object detector has a better impact on the density estimation than the dense SIFT descriptors in 7 of 8 cases (the one exception stems for $L_1$ based regularization, which is in general unstable). Using both features increases the accuracy. It is also interesting that the two regularizations yield similar results, even though the learned weights are very different. Overall, the $L_1$ regularization tends towards a \textit{zero}-solution, \ie setting many weights to zero, while the Tikhonov regularization populates the weights a lot \textit{smoother} (this is a property of the Tikhonov regularization, as it improves the condition of the problem and enables a more stable numerical solution). This aspect seems not important for the learning set, however it changes the performance in the testing phase. If \eg we have a slight motion blur in one of the images, the according $L_1$ weights drop to zero, while the Tikhonov weights do not. For the \textit{Donauinsel} scenario the Tikhonov based regularization yields a lower accuracy than $L_1$ in case of dense SIFTs. We assume that the low number of learning samples and the unfavorable mapping of discretized SIFT values to the real occurrence of persons (the stage rack contains many vertical structures, \ie the same features of a person) yield a bad condition of the equation system and therefore the solution tends to a local minimum instead of the global one.
%
%\reffig{figure:learned_weights} shows the described behaviors. The first 256 features represent the quantized SIFT keys and the second 256 the discretized object detection scores.
%
While learning based in $L_1$ regularization picks a few SIFT keys and a few object detection scores (only 10), the Tikhonov based learning takes more SIFTs and a logical weight distribution of the object detector (in total 453). Where logical means that the learned weights are dependent on the object detector confidences.
%
%
%Tikhonov much slower
%
\begin{table}[!h]
\vspace{\vspacetablePre}
\centering
%\scriptsize
\footnotesize{
\begin{tabular}{|c|c|c||c|c|}
\hline \textit{Lakeside} & \multicolumn{2}{|c||}{training} & \multicolumn{2}{|c|}{testing} \\
                & \hphantom{MM} $L_1$ \hphantom{MM} & Tikhonov & \hphantom{MM} $L_1$ \hphantom{MM} & Tikhonov\\
\hline
\hline object detector & 4.7 (1.8\%) & 4.75 (1.8\%) & 13.3 (4.8\%)& 10.6 (3.8\%) \\
\hline dense SIFT      & 7.0 (2.7\%) & 6.7 (2.5\%) & 11.2 (4.0\%)& 11.1 (4.0\%) \\
\hline both            & 4.5 (1.7\%) & 4.4 (1.7\%) & 10.8 (3.9\%) & 10.0 (3.6\%) \\
\hline
\multicolumn{5}{c}{}\\% [-1.5ex]
%\end{tabular}
%
%\vspace{1mm}
%
%\begin{tabular}{|c|c|c||c|c|}
\hline \textit{Donauinsel} & \multicolumn{2}{|c||}{training} & \multicolumn{2}{|c|}{testing} \\
                & \hphantom{MM} $L_1$ \hphantom{MM} & Tikhonov & \hphantom{MM} $L_1$ \hphantom{MM} & Tikhonov\\
\hline
\hline object detector & 7.1 (5.3\%) &  7.0 (5.2\%) & 12.7 (9.0\%)  & 10.0 (7.1\%) \\
\hline dense SIFT      & 7.0 (5.2\%) & 10.3 (7.7\%) & 15.9 (11.3\%) & 18.0 (12.8\%) \\
\hline both            & 7.1 (5.3\%) &  5.6 (4.2\%) & 11.9 (8.4\%)  & 12.1 (8.6\%) \\
\hline
\end{tabular}
}
\vspace{\vspacetab}
\caption[]{Accuracy of density learning and testing. Given are the average errors of the total human count and the percental error over the training and test images, for two regularization options and different image features.} \label{table:accuracy}
\end{table}
%
%\begin{figure}[h]
%\vspace{\vspacefigPre}
%\centering
%\begin{tabular}{cc}
%\resizebox*{\sizefigTwo\columnwidth}{!}{\includegraphics{_weights_1.png}} & \resizebox*{\sizefigTwo\columnwidth}{!}{\includegraphics{_weights_2.png}}
%\end{tabular}
%\vspace{\vspacefig}
%\caption[]{Density learning: The learned weights for the combined features (dense SIFT and object detector) are given for (left) $L_1$ regularization and (right) Tikhonov regularization. While the %solution of (left) contains many \textit{zero}-weights (502), the solution of (right) contains significantly less (59).} \label{figure:learned_weights}
%\vspace{\vspacefigPost}
%\end{figure}
%
\mysection{Testing} The accuracy of the density estimation is given in \reftbl{table:accuracy}. Like in the training phase the Tikhonov regularization yields slightly higher accuracies than the $L_1$ one. On average the mean person counting error is $4\%$ of the \textit{Lakeside} and $9\%$ for the \textit{Donauinsel} data set. \reffig{figure:counts} shows the estimated person count of \textit{Lakeside} with superimposed manually measured ground truth. Both resulting curves are similar however the Tikhonov regularization creates a smoother result. Experimentally we can prove this assumption by taking a look at the temporal smoothness of the estimated person count. The standard deviation of per frame differences of the estimated count is $4.4$ for $L_1$ regularization and $3.8$ for Tikhonov regularization (for \textit{Lakeside} and when using both feature sets). Obviously, a lower number represents a more realistic setting, as the number of persons in two adjacent frames should not vary much. When taking a close look to \reffig{figure:counts} a rather huge error is visible towards the end of the sequence (image number 6500 to 6700). The reason for this issue are strong winds causing camera shaking and therefore a motion blur in the images. Consequently, the extracted features are different to the learned weights resulting in a lower human density estimate. 
%
%\reffig{figure:boxplots} visualizes the statistics of absolute errors in terms of box plots for different features and regularizations. It could be seen that using all features and the Tikhonov regularization results in the smallest standard deviation and to no gross outliers.
%
% (std: 7.2, 5.5)
% (std: 3.2, 3.1)
% (std: 4.4, 3.8)
%
%
%\begin{figure}[h!]
%\centering
%\begin{tabular}{c}
%\resizebox*{\sizefigOne\columnwidth}{!}{\includegraphics{_boxplot_both.png}}\\
%\end{tabular}
%\vspace{\vspacefig}
%\caption[]{Person counting uncertainty: Box plots for the the \textit{Lakeside} scenario.} \label{figure:boxplots}
%\end{figure}
% ---------------------------------------------------------------------------------------------------------------------
%\section{Discussion}
%
\section{Conclusion}
In this work we presented a method for people counting and crowd monitoring from airborne imagery. The estimated parameters from a given video stream were human count and human density and motion for each pixel. This information was geo-referenced into a world coordinate system. Overall, the estimated human counts were highly accurate with resulting 4\% and 9\% count error for the two presented scenarios, which could be reached by employing a custom tailored object detector instead of simple images features amongst other implementation details. The proposed framework is therefore higly important for security applications.
\mysection{Outlook} Currently, the framework is optimized for oblique views and thus it will not yield reasonable accuracies when \eg employing nadir images. We envision to train the system on several viewing conditions, where the object detector should also be custom tailored (like a detector for head and shoulders for oblique views and a blob-like detector for nadir views). The viewing condition itself can be derived from the airplane's geo-sensors. When extracting the human densities the system is able to choose from the learned models according to the viewing parameters. Of course it would also be of interest to test different features and detectors on the accuracy and various regularizations for minimizing the \textit{MESA} distance in the machine learning approach.
\subsubsection*{Acknowledgments.}
This research has been funded by the Ministry of Austria for Transport, Innovation and Technology (bmvit) within the security research program KIRAS: Project 821733 \qm{EVIVA - Airborne based monitoring and analysis system for event protection using video based behavior analysis} and Project 832353 \qm{EN MASSE - System for multi-sensor crowd monitoring for real time visualization of a common operational picture (COP) and short-time forecast}.

\bibliography{2013_oagm}

\begin{thebibliography}{10}

\bibitem{butenuth2011integrating}
M.~Butenuth, F.~Burkert, F.~Schmidt, S.~Hinz, D.~Hartmann, A.~Kneidl,
  A.~Borrmann, and B.~Sirma{\c{c}}ek.
\newblock Integrating pedestrian simulation tracking and event detection for
  crowd analysis.
\newblock In {\em ICCV Workshops}, pages 150--157, 2011.

\bibitem{chan2008privacy}
A.B. Chan, Z.-S.J. Liang, and N.~Vasconcelos.
\newblock Privacy preserving crowd monitoring: {C}ounting people without people
  models or tracking.
\newblock In {\em CVPR}, pages 1--7, 2008.

\bibitem{everingham2009the}
M.~Everingham, L.~Van~Gool, C.~K.~I. Williams, J.~Winn, and A.~Zisserman.
\newblock The {PASCAL} {V}isual {O}bject {C}lasses {C}hallenge 2009 {(VOC2009)}
  {R}esults.

\bibitem{felzenszwalb2010object}
P.~F. Felzenszwalb, R.~B. Girshick, D.~McAllester, and D.~Ramanan.
\newblock Object detection with discriminatively trained part based models.
\newblock {\em IEEE Transactions on Pattern Analysis and Machine Intelligence},
  32(9):1627--1645, 2010.

\bibitem{helbing2007dynamics}
D.~Helbing, A.~Johansson, and H.Z. Al-Abideen.
\newblock Dynamics of crowd disasters: {A}n empirical study.
\newblock {\em Physical Review E}, 75:046109, 2007.

\bibitem{kong2006a}
D.~Kong, D.~Gray, and H.~Tao.
\newblock A viewpoint invariant approach for crowd counting.
\newblock In {\em ICPR}, volume~3, pages 1187--1190, 2006.

\bibitem{kraus2007photogrammetry}
K.~Kraus and I.~A. Harley.
\newblock {\em Photogrammetry: Geometry from Images and Laser Scans}, volume~1.
\newblock de Gruyter Textbook, 2nd edition, 2007.

\bibitem{lempitsky2010learning}
V.~Lempitsky and A.~Zisserman.
\newblock Learning to count objects in images.
\newblock In J.~Lafferty, C.~K.~I. Williams, J.~Shawe-Taylor, R.S. Zemel, and
  A.~Culotta, editors, {\em NIPS}, number~23, pages 1324--1332. 2010.

\bibitem{lowe2004distinctive}
D.~G. Lowe.
\newblock Distinctive image features from scale-invariant keypoints.
\newblock {\em International Journal of Computer Vision}, 60(2):91--110, 2004.

\bibitem{sharp2008implementing}
T.~Sharp.
\newblock Implementing decision trees and forests on a gpu.
\newblock In {\em ECCV (Part IV)}, pages 595--608, 2008.

\bibitem{sirmacek2011automatic}
B.~Sirma{\c{c}}ek and P.~Reinartz.
\newblock Automatic crowd density and motion analysis in airborne image
  sequences based on a probabilistic framework.
\newblock In {\em ICCV Workshops}, pages 898--905, 2011.

\bibitem{tikhonov1977solutions}
A.N. Tikhonov and V.Y.Arsenin.
\newblock {\em Solutions of Ill Posed Problems}.
\newblock WH Winston, Washington, DC, 1977.

\bibitem{vedaldi08vlfeat}
A.~Vedaldi and B.~Fulkerson.
\newblock {VLFeat}: An open and portable library of computer vision algorithms.
\newblock {\url{http://www.vlfeat.org}}, 2008.

\bibitem{viola01rapid}
P.~Viola and M.~Jones.
\newblock Rapid object detection using a boosted cascade of simple features.
\newblock In {\em CVPR}, volume~I, pages 511--518, 2001.

\bibitem{zach2007a}
C.~Zach, T.~Pock, and H.~Bischof.
\newblock A duality based approach for realtime tv-l1 optical flow.
\newblock In {\em Pattern Recognition (Proc. DAGM)}, pages 214--223,
  Heidelberg, Germany, 2007.

\end{thebibliography}
%\balance

\end{document}